\documentclass[conference]{IEEEtran}
\usepackage{cite}

\ifCLASSINFOpdf
  \usepackage[pdftex]{graphicx}
\else
\fi
\usepackage{amsmath}
\usepackage{algorithmic}
\usepackage{algorithm}
\ifCLASSOPTIONcompsoc
  \usepackage[caption=false,font=normalsize,labelfont=sf,textfont=sf]{subfig}
\else
  \usepackage[caption=false,font=footnotesize]{subfig}
\fi

\newcommand{\argmin}{\operatornamewithlimits{argmin}}
\newcommand{\argmax}{\operatornamewithlimits{argmax}}

\begin{document}
%
\title{On Reward Function for Survival}

\author{\IEEEauthorblockN{Naoto~Yoshida}
\IEEEauthorblockA{Tohoku University, Sendai, Japan\\ E-mail: naotoyoshida@pfsl.mech.thoku.ac.jp}
}


%


\maketitle

\begin{abstract}
Obtaining a survival strategy (policy) is one of the fundamental problems of biological agents. In this paper, we generalize the formulation of previous research related to the survival of an agent and we formulate the survival problem as a maximization of the multi-step survival probability in future time steps. We introduce a method for converting the maximization of multi-step survival probability into a classical reinforcement learning problem. Using this conversion, the reward function (negative temporal cost function) is expressed as the log of the temporal survival probability. And we show that the objective function of the reinforcement learning in this sense is proportional to the variational lower bound of the original problem. Finally, We empirically demonstrate that the agent learns survival behavior by using the reward function introduced in this paper.
\end{abstract}


%
\IEEEpeerreviewmaketitle

\section{Preliminaries}
\IEEEPARstart{S}{urvival} strategies are essential for biological agents. Many researchers have developed various types of survival agents since the early days of artificial intelligence. Ashby developed Homeostat, which dynamically stabilizes the state of the machine\cite{ashby1960design},  and Walter developed simple robotic agents that can explore the environment of a room and automatically recharge their batteries at a recharging station \cite{walter1953living}. Toda discussed the survival problem of artificial agents in the natural environment \cite{toda1962design,toda1982man}. He speculated about the functional requirements for an autonomous survival agent based on decision theory. Based on Toda's works, Pfeifer and Scheier pointed out the importance of `complete' autonomous agents and research on embodied cognitive science \cite{pfeifer1999understanding}. In this sense, they developed a simple self-sufficient autonomous robot. McFarland and B\"{o}sser discussed the autonomous agent from the perspective of research on animal behavior \cite{mcfarland1993intelligent}. 
They suggested several requirements for intelligent agents through comparison of the market economy with natural selection, and they also developed simple robots that were self-sufficient, autonomous agents \cite{mcfarland1997basic}.
Also, Lin, in a simulation study, compared several reinforcement learning (RL) architectures for a complex survival task in a non-Markovian environment \cite{lin1992self}. Sibly and McFarland introduced the state space approach in ethology and suggested that animal behaviors are the consequence of optimal control with respect to the cost function given by the fitness function \cite{sibly1976fitness}. Keramati and Gutkin suggested a similar perspective, but they also suggested changes in the distance between the current homeostatic state and the optimal, desired state as the reward function of RL \cite{keramati2011reinforcement}. Konidaris and Barto developed RL architecture that automatically balances multiple required nutrients (protein, fat, water, etc.) through tuning of the reward function, which depends on the agent's homeostatic state \cite{konidaris2006adaptive}. They tested the architecture in a simulation experiment.

Ogata and Sugino developed a real robot agent intended for survival \cite{ogata1997emergence}. Their robot evaluates the sensor signals (motor temperature, battery level, etc.) in real time and learns to avoid undesirable stimuli. Doya and Uchibe developed robots called  ``Cyber rodents'' intended for the study of learning agents for survival and evolutionary robotics \cite{doya2005cyber}. Cyber rodents can recharge their batteries by touching special battery packs in the environment. The wireless communication modules of robots enable the software evolution of control algorithms  \cite{elfwing2005biologically}.

Reward stimuli have been introduced in most of the  preceding research studies, with the reward function being necessary for the RL paradigm. However, almost all of the previous studies on survival adopted hand-crafted reward functions that do not guarantee the intended behaviors, which are the survival strategies in this case. The reward and objective function given by the designer may work well in simple RL tasks. However, for more complex tasks and life-long learning settings in which agents must learn for days and months with a large amount of data, a badly hand-crafted reward function may ultimately have serious problems in terms of  system performance.   

Based on such considerations, we first focus on the mathematical formulation of the classical survival problem as a maximization problem of the multi-step survival probability. To solve this problem, we introduce an iterative model-based method for maximization of the objective function using an expectation-maximization (EM) algorithm with variational approximations. Surprisingly, after the M-step, the negative free-energy function (variational lower bound) in our algorithm is identical in form to the classical RL objective function with a specific reward function. Therefore, it can be maximized by classical model-free RL algorithms. Our contributions are i) a probabilistic formulation of the classical survival problem, ii) a suggested RL approach to solve this problem and iii) a demonstration that the maximization of multi-step survival probability through RL algorithms is identical to the maximization of the log of that probability from the variational lower bound.

\subsection{Objective Functions of Reinforcement Learning Problem}
Reinforcement learning (RL) is the field of research that constructs learning agents that obtain an optimal policy through interactions with the environment. We first introduce the general form of the objective function in the POMDP (partially observable Markov decision process) model. For the sake of simplicity, we restrict the discussion to a finite set of actions, states, and observations. 

Many realistic environments for the agent are known to be modeled by the partially observable Markov decision process (POMDP) \cite{kaelbling1996reinforcement}. The POMDP model consists of the state set ${\cal S}$, the action set ${\cal A}$, the observation set ${\cal O}$, the transition probability to state $s'  \in {\cal S}$ given a state $s  \in {\cal S}$ and an action $a  \in {\cal A}$ as $P(s'|s,a)$, the observation probability $P(o|s)$, and the reward function $r(s,a)$. 

At each time step $t$, the agent receives an observation $o_t$ from the state $s_t$ by the observation probability and replies with an action $a_t$. Then, the state of the environment changes to $s_{t+1}$ and the agent receives a reward $r_t = r(s_t,a_t)$. In the POMDP setting, since the agent can not gain access to the true state $s \in {\cal S}$ of the environment, the agent needs to infer the current true state from a sequence of observations and actions. We call the sequence of observations and actions a {\bf history} $h_t = \{o_0, a_0, o_1, a_1, \dots, o_{t-1}, a_{t-1}, o_t\}$. Using the history, how the agent acts in the environment can be expressed by a probability $\pi(a|h)$ called the {\bf policy}. In the POMDP model, the probability of generating the $T$-step {\bf trajectory} $\tau_T = \{s_0, o_0, a_0, s_1, o_1, a_1, \dots, s_{T-1}, o_{T-1}, a_{T-1}, s_{T}, o_{T}\}$ is  
\begin{eqnarray*}
P(\tau_T| \pi) &=& P(o_0|s_0) P(s_0) \\
&&\times  \prod^{T-1}_{t=0}P(o_{t+1}|s_{t+1}) P(s_{t+1}|s_t, a_t) \pi(a_t|h_t).
\end{eqnarray*}

Therefore, the expectation of the $T$-step average reward is given by 
\begin{eqnarray*}
J_T(\pi) = \sum_{\tau_T} P(\tau_T| \pi) \Bigl[\frac{1}{T} \sum^{T-1}_{t=0} r_t \Bigr].
\end{eqnarray*}
 We denote the limit $T\rightarrow \infty$ by $J(\pi) = \lim_{T\rightarrow \infty}J_T(\pi)$. This $J_T(\pi)$ or $J(\pi)$ (or the product with some constant) is the typical objective function in the reinforcement learning literature\footnote{Even though many studies derive the algorithm from another objective function like $\sum^{T-1}_{t=0} \gamma^t r_t$, the performance of the algorithm is usually evaluated by the total or the average reward criterion.}. The objective of reinforcement learning is to find the optimal policy $\pi^*$ that maximally achieves the objective function, defined above, through interactions with the environment.

\section{Survival Problem}
\begin{figure}[!t]
\centering
\includegraphics[width=4cm]{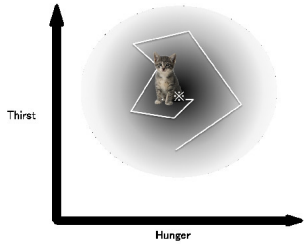}
\caption{
The State Space Model: The figure shows the relationships between the physiological state and the viability zone. In this figure, the viability zone depends only on the two continuous variables, the  ``energy level" and ``water level" for the sake of simplicity. The star represents the position of the optimal physiological state. The agent can remain alive in the viability zone (darker area) and should be able to recover from a position separated from this area.
}
\label{fig_statespace}
\end{figure}

In this section, we formulate the survival problem from the models of an animal proposed by Ashby \cite{ashby1960design} and a similar idea suggested by Sibly and McFarland \cite{sibly1976fitness} and McFarland and B\"{o}user \cite{mcfarland1993intelligent} from the view point of ethology. In their model, an animal has several variables that are observed by the animal and have some importance for sustaining life (for example, the water level and the energy level in the body, as shown in Figure \ref{fig_statespace}). We call these variables the `physiological state'. On the other hand, an agent also has a `perceptual state', which is the perception of the environmental stimuli (vision, touch, etc.). The combined physiological state and perceptual state, which may be represented in the animal's brain, is called the  `motivational state' \cite{mcfarland1993intelligent}\footnote{The terms `physiological state', `perceptual state' and `motivational state' may be misleading in this paper, because these ``states'' do not necessarily follow Markovian dynamics.}.
The animal has one compact manifold, which is defined in the physiological state space. This manifold is called the viability zone \cite{meyer1991simulation}, and we define the state of the animal as `Alive' when the current physiological state is in the manifold. The adaptive behavior is expressed as an optimal process that steers the physiological state toward the optimal state using the observed data from outside and inside the body.

\subsection{Formulation of the Survival Problem}
\begin{figure}[!t]
\centering
\includegraphics[width=5cm]{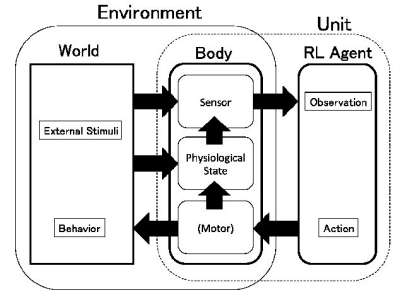}
\caption{The settings in the survival problem with an RL agent. A unit of the agent consists of a body and a RL agent. The RL agent interacts with the world through the body.  The world receives the motor outputs from the body, returns the (external) stimulus, and changes the physiological state of the body. The body receives an external stimulus through the sensors while monitoring the physiological state through other sensors. Then, the body sends an observation to the RL agent. The RL agent receives the observation and responds with an action.
}
\label{fig_settings}
\end{figure}

The assumptions of the problem are as shown in Figure \ref{fig_settings}. Similar settings have been suggested elsewhere \cite{ashby1960design, mcfarland1981quantitative, canamero1997modeling, barto2004intrinsically, konidaris2006adaptive}. A unit of the agent consists of an RL agent and a body, and the RL agent interacts with the world through the body. Because the sensors may not perfectly determine the current situation of the world and the physiological state of the body, the sensor may exhibit only partial observability. Since sensing of the physiological state is a process inside of the body, we may be able to assume that there is no information loss with this sensing.

We now formalize the survival problem as an optimization problem. We mostly follow the usual definition of the dynamics in the POMDP model explained in the previous section. Like the POMDP model, the agent interacts with the environment. At the state $s_t \in {\cal S}$, the agent receives the current observation $o_t \in {\cal O}$ with probability $P(o|s)$. Then, the agent takes an action $a_t \in {\cal A}$ following some policy $\pi$; the state then changes at the next time step following the probability $P(s_{t+1}|s_t,a_t)$. Because  $o_t$ is the observation which consists of the temporal stimulus to the agent at the time step $t$, we can understand that the observation in POMDP is a generalization of the motivational state. In this definition, we do not explicitly separate the observation caused by external stimuli (the perceptual state; vision, touch, etc.) from the observation caused by internal stimuli (the observed physiological state; energy level, water level, etc.).

In order to formulate the survival problem, we introduce a binary signal $A_t$ instead of the reward, which represents the ``alive flag" of the agent at times step $t = 0, 1, 2, 3 \cdots$. $A_t = 1$ represents that the agent is `Alive' at the time step $t$, and $A_t = 0$ represents `Dead'. To generalize the problem, we soften the definition of the boundary of the viability zone by introducing the temporal survival probability $P(A_{t+1} = 1 | s_t) > 0$. Because the survival probability is ultimately caused by the physiological state of the agent (animal), we assume that the survival probability is only dependent on the current state. However, the following discussion is directly applicable for other definitions of the temporal survival probability, including $P(A_{t+1} = 1 | o_t)$, $P(A_{t+1} = 1 | s_t, a_t)$, and so on. Also, we assume $A_0 = 1$, and this probability is known for the agent unit (either in the body or RL agent). 

A natural interpretation of `survival' is to stay alive as long as possible in the environment, so the agent requires policy $\pi$ that realizes the signal sequence $A_t = 1$ $(t=0,1,\dots)$. Using these definitions, the multi-step survival probability given a policy $\pi$ is expressed by a joint probability
 \begin{eqnarray}
P({\bar A}_T| \pi),
\label{eq:suv}
\end{eqnarray}
where ${\bar A}_T$ denotes the sequence $\{A_0=1, A_1=1, \dots, A_{T+1}=1\}$. We then define the objective of the agent for the survival problem as the maximization of the probability defined by (\ref{eq:suv}).

\subsection{Maximization of Survival Probability by Variational Method}
In the following discussion, we show that the maximization of the objective function (\ref{eq:suv}) through maximization of the variational bound can be reduced to maximization of the conventional objective function of RL. 

First, we discuss the situation of the planning problem, in which we search for the optimal policy given the true transition probability $P(s'|s,a)$, the observation probability $P(o|s)$, and the temporal survival probability $P(A|s)$. The logarithm of the objective function (\ref{eq:suv}) can be transformed by introducing the arbitrary probability distribution $Q(\tau)$ on the $T$-step trajectory $\tau$ as 
\begin{eqnarray*}
\nonumber \log P({\bar A}_T|\pi) &=& \sum_\tau Q(\tau) \log P(\bar{A}_T|\pi)\\
\nonumber &=&\sum_\tau Q(\tau) \log \frac{ P({\bar A}_T, \tau|\pi)}{P(\tau| {\bar A}_T, \pi)}\frac{Q(\tau)}{Q(\tau)}\\
\nonumber &=& \sum_\tau Q(\tau) \Bigl( \log P(\bar{A}_T|\tau) - \log \frac{Q(\tau)}{P(\tau|\pi)} \Bigr)\\
&&\ \ \ \ \ + \sum_\tau Q(\tau) \log \frac{Q(\tau)}{P(\tau| \bar{A}_T, \pi)}\\
&=& -F(Q(\cdot), P(\cdot|\pi)) + \rm{KL}(Q(\cdot) || P(\cdot| \bar{A}_T, \pi)).
\label{eq:log_fe}
\end{eqnarray*}
The relationship $P(\bar{A}_T|\pi)P(\tau| \bar{A}_T, \pi) = P(\bar{A}_T, \tau|\pi)$ was used at the second equality. $P(\tau| \bar{A}_T, \pi)$ is the posterior of $P(\tau|\pi)$ given ${\bar A}$, which is $P(\tau| \bar{A}_T, \pi)=\frac{P(\bar{A}_T|\tau)P(\tau|\pi)}{\sum_{\tau'}P(\bar{A}_T|\tau')P(\tau'|\pi)}$ from Bayes's theorem, and $P(\bar{A}_T, \tau | \pi) = P(\bar{A}_T| \tau)P(\tau | \pi)$ was used at the third equality. The first term on the RHS in the last row is 
\begin{eqnarray*}
-F(Q(\cdot), P(\cdot|\pi)) = \sum_\tau Q(\tau) \Bigl( \log P(\bar{A}_T|\tau) - \log \frac{Q(\tau)}{P(\tau|\pi)} \Bigr),
\end{eqnarray*}
and $F$ is called the free energy from the analogy with the statistical mechanics. If there is some restriction on the distribution $Q(\tau)$ (for example, $Q(\tau)$ is given by a specific class of the probability distribution),  $F$ is called the variational free energy. ${\rm KL}(Q(\cdot)||P(\cdot))$ is the Kullback-Leibler (KL) divergence ${\rm KL}(Q(\cdot)||P(\cdot)) = \sum_\tau Q(\tau) \log \frac{Q(\tau)}{P(\tau)}$. The KL divergence is known to be non-negative and zero only if the two probability distributions $P$ and $Q$ are equivalent. Because of the non-negativity of the KL divergence and the above equality, the log-probability $\log P(\bar{A}_T|\pi)$ is lower bounded by $-F$. Hence, this value is termed the variational (lower) bound.

Also, from the equality above
\begin{eqnarray*}
\log P(\bar{A}_T|\pi) + F(Q(\cdot), P(\cdot|\pi)) = \rm{KL}(Q(\cdot) || P(\cdot| \bar{r}_T, \pi)),
\end{eqnarray*}
and there is a relationship
\begin{eqnarray*}
\argmin_Q [F(Q(\cdot), P(\cdot|\pi))]= \argmin_Q [\rm{KL}(Q(\cdot) || P(\cdot| \bar{A}_T, \pi))]
\end{eqnarray*}
because $\log P(\bar{A}_T|\pi)$ is not a function of $Q$.

In the maximization problem of the likelihood $\log P(\bar{A}_T|\pi)$, the method that introduces the restricted class of $Q(\tau)$ and maximizes the variational bound $-F$ is known to be the variational method and it is widely used in machine learning \cite{jordan1999introduction}. The probability distribution of the $T$-step trajectory $\tau = \{s_0, o_0, a_0, s_1, o_1, a_1, \dots, s_{T-1}, o_{T-1}, a_{T-1}, s_{T}, o_{T}\}$ given a policy $\pi$ is
\begin{eqnarray*}
P(\tau|\pi)&=&P(o_0|s_0) P(s_0) \\
&&\times \prod^{T-1}_{t=0}P(o_{t+1}|s_{t+1}) P(s_{t+1}|s_t, a_t) \pi(a_t|h_t).
\end{eqnarray*}
And we restrict the distribution $Q(\tau)$ with arbitrary policy $\pi_Q(a|h)$ to
\begin{eqnarray*}
Q(\tau|\pi_Q)&=& P(o_0|s_0) P(s_0) \\
&&\times\prod^{T-1}_{t=0}P(o_{t+1}|s_{t+1}) P(s_{t+1}|s_t, a_t) \pi_Q(a_t|h_t).
\end{eqnarray*}

\begin{algorithm}[t]
\algsetup{linenosize=\small, linenodelimiter=.}
\caption{}
\begin{algorithmic}[1]
\STATE Set $k=0$ and an arbitrary policy $\pi^0$.
\STATE (E-step) Obtain $\pi^k_Q$ by optimization 
\begin{eqnarray*}
\pi^k_Q &=&  \argmin_{\pi_Q} [\rm{KL}(Q(\cdot|\pi_Q) || P(\cdot| \bar{A}_T; \pi^k))]\\
 &=& \argmax_{\pi_Q} [-F(Q(\cdot|\pi_Q), P(\cdot|\pi^k))].
\end{eqnarray*}
\STATE (M-step) Update $\pi$ by using $\pi^k_Q$. That is
\begin{eqnarray*}
\pi^{k+1} &=& \argmax_{\pi} [-F(Q(\cdot|\pi_Q^k), P(\cdot|\pi))]\\
 &=& \pi^k_Q.
\end{eqnarray*}
\IF {$\pi^{k+1}$ is converged,}
\STATE return $\pi^{k+1}$
\ELSE
\STATE $k\leftarrow k+1$ and go to E-step.
\ENDIF
\end{algorithmic}
\end{algorithm}

By using these distributions, we introduce the EM algorithm as Algorithm 1. The maximization in the M-step is simply the replacement of $\pi^k$ by $\pi^k_Q$ from the equality
\begin{eqnarray*}
 &&\argmax_{\pi}[-F(Q(\cdot|\pi_Q), P(\cdot|\pi))]\\
 &=& \argmax_{\pi}\Bigl[\sum_\tau Q(\tau|\pi_Q) \Bigl( \log P(\bar{A}_T|\tau) - \log \frac{Q(\tau|\pi_Q)}{P(\tau|\pi)} \Bigr)\Bigr]\\
&=&\argmin_{\pi} \Bigl[\sum_\tau Q(\tau|\pi_Q) \log \frac{Q(\tau|\pi_Q)}{P(\tau|\pi)}\Bigr]\\
&=& \argmin_{\pi} [{\rm KL}(Q(\cdot|\pi_Q)||P(\cdot|\pi))]\\
&=&\pi_Q.
\end{eqnarray*}

Because of the restriction on $Q(\tau|\pi_Q)$, the minimization of the KL divergence in the E-step is a variational sense and may not be zero. If the environment is MDP and we assume that $\pi_Q$ is a Markov policy, Rawlik et al. derived the analytical solution of the E-step and it is given by the softmax policy $\pi_Q(a|s) = \exp\{\Psi(s,a)\}/Z$, where $\Psi(s,a)$ is some energy function and $Z$ is the normalization term \cite{rawlik2013stochastic}.

In POMDP, on the other hand, no analytical solution to the E-step is known. To tackle this problem, we may parametrize the policies $\pi$ and $\pi_Q$ by parameters $\theta$, $\phi$ as $\pi_\theta$, $\pi_\phi$. And then, we assume $\theta = \phi \Rightarrow \pi_\theta = \pi_\phi$. Therefore $\theta = \phi \Rightarrow P(\tau|\pi_\theta) = Q(\tau|\pi_\phi)$ by definition. The variational method that introduces the parametrized variational distribution $Q_\phi$ and maximizes the variational bound $ -F(\phi,\theta) := -F(Q(\cdot|\pi_\phi), P(\cdot|\pi_\theta)) $ by gradient methods are well known in the neural computing community \cite{dayan1996varieties, ranganath2013black, kingma2013auto, mnih2014neural}. Following this idea, we replace the E-step and M-step in Algorithm 1 by
\begin{eqnarray}
\phi^k = \argmax_{\phi} [-F(\phi, \theta^k)]
\label{algo2_1}
\end{eqnarray}
and
\begin{eqnarray}
\theta^{k+1} = \argmax_{\theta} [-F(\phi^k, \theta)] = \phi^k.
\label{algo2_2}
\end{eqnarray}

If each stage of the algorithm is performed exactly, a monotonic increase of the variational bound
\begin{eqnarray*}
-F(\phi^0,\theta^{1})\leq -F(\phi^1,\theta^{2}) \leq -F(\phi^2,\theta^{3}) \leq \dots
\end{eqnarray*}
after the M-step is guaranteed. Moreover, there is a relationship after the M-step
\begin{eqnarray*}
-F(\phi^k,\theta^{k+1})&=&-F(\theta^{k+1}, \theta^{k+1})\\
&=& \sum_\tau P(\tau| \pi_{\theta^{k+1}}) \log P(\bar{A}_T|\tau)\\
&=& \sum_\tau P(\tau| \pi_{\theta^{k+1}}) \Bigl[\sum^{T-1}_{t=0} \log P(A_{t+1}=1|s_t)\Bigr]\\
&=& T J_T(\pi_{\theta^{k+1}})
\end{eqnarray*}
in which $J_T(\pi)$ denotes the objective function of the reinforcement learning with respect to the reward function $r_t = \log P(A_{t+1}=1|s_t)$. Here, the equality $\log P(\bar{A}_T|\tau) = \sum^{T-1}_{t=0} \log P(A_{t+1}=1|s_t)$ is used at the third equality. Because $T$ is a constant, an increase of the variational bound is equivalent to the increase of $J_T(\pi_{\theta^{k+1}})$. Therefore, from the discussion above, the maximization of the log-form of the objective function $\log P(\bar{A}_T|\pi_\theta)$ through the variational bound $-F(\phi,\theta)$ is reduced to the maximization of $J_T(\pi_{\theta})$ with respect to the parameter of the agent $\theta$. 

\subsection{Solving the Survival Problem by Reinforcement Learning Algorithms}
Now we consider the survival problem in the reinforcement learning setting; that is, the maximization of the log of the objective function $\log P(\bar{A}_T|\pi)$ while the agent cannot access the true environment model $P(o|s)$, $P(s'|s,a)$. In this setting, we can not perform the iterative algorithms described above. However, from the discussion of the second algorithm (equation \ref{algo2_1}, \ref{algo2_2}), the variational bound is proportional to $J_T(\pi_{\theta})$ after each M-step. Then, in order to maximize the variational bound, we can take a direct maximization of $J_T(\pi_{\theta})$ with respect to $\theta$, instead of the exact execution of the iterative algorithm. Because $J_T(\pi_{\theta})$ with reward function 
\begin{equation*}
r_t = \log P(A_{t+1}=1|s_t)
\end{equation*}
is the conventional objective function of the reinforcement learning paradigm, we can apply the RL algorithms to the maximization of the survival probability.

\section{Experiment}
In the experiment, we verify the reward setting by evaluating the finite horizon survival probability in the simple grid world domain. 

\paragraph{Environment}
\begin{figure}[!t]
\centering
\includegraphics[width=3cm]{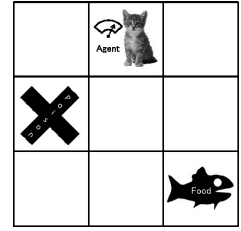}
\caption{The grid world environment. There are two objects, {\bf A} and {\bf B}, but the agent initially does not know which corresponds to the ``food" object. Further details of the environment are explained in the main text.}
\label{environment}
\end{figure}

The environment consists of a 3 x 3 grid world (Figure \ref{environment}). The agent selects an action at each time step from UP, DOWN, RIGHT, LEFT and EAT. When the agent takes the action UP, DOWN, RIGHT or LEFT and if the wall is not in that direction, the agent moves one step in the selected direction. Otherwise, the agent stays at the current position. In the environment, there are two types of objects, {\bf A} and {\bf B}, at uniformly random positions, such that the two objects never overlap. The position of the objects changes if the agent selects the EAT action at the corresponding position. Also, the position of {\bf B} may randomly change at every time step with a probability of 0.01.   

The agent has a continuous battery level $E \in [0, 100]$ that decreases by $1\%$ at each time step. The battery level is recharged $+5$ if the agent selects EAT at the position of object {\bf A}. Therefore, the object {\bf A} corresponds to the food object.

The temporal survival probability of the agent when $A_{t} = 1$ is defined as $P(A_{t+1} =1 | s_t) = P(A_{t+1} =1 | E_t, C_t) = f(E_t) g(C_t)$
where $E_t$ is the battery level at time $t = 1, 2, 3, \dots$, $C_t \in \{0, 1\}$ is the flag bit whether the agent ate the object {\bf B} ($C = 1$) or not ($C = 0$).  $f$ and $g$ are defined as
\begin{eqnarray*}
 f(E_t)  = \exp \Bigl\{ - \frac{(E_t - 60)^2}{1000} \Bigr\}
\end{eqnarray*}
and
\begin{eqnarray*}
 g(C_t) 
 = \begin{cases}
    0.5 & (C_t = 1) \\
    1    & (C_t = 0).
  \end{cases}
\end{eqnarray*}

The observation of the agent is defined by the set $o = \{x, p^A, p^B, c, \hat E\}$, where $x$ is the position of the agent, $p^A$ is the position of object {\bf A}, $p^B$ is the position of object {\bf B}, $c_t$ is the type of object that the agent EATs ($c=1$ for nothing, $c=2$ for {\bf A}, and $c=3$ for {\bf B}), and $\hat E$ is the discrete state of the current battery level. 
In this experiment, we discretized the continuous battery level $[0, 100]$ into 20 discrete regions, and $\hat E$ receives the class of the corresponding region of the current battery level $E$. Therefore, this environment is a simple POMDP setting because of the discretization of the battery level. In order to survive in this environment, the agent must take the food ({\bf A}), avoid the poisonous object ({\bf B}) and regulate its energy level ($E$).
Even though this might be an over simplified model of the biological agent, this kind of situation will occur everywhere in the life of animals. Importantly, in this setting, agents initially do not know which object information ($p^A$, $p^B$) corresponds to ``food" or ``poison".
Then the agent has to associate these objects with changes in the homeostatic values ($E$ and $c$). Also, agents never receive positive rewards when they take food and the reward values for food-capture depend on the agents' battery level. Therefore, this experiment is fundamentally different from task-oriented problems like ``food capturing".

\paragraph{Agent Settings}
The Sarsa($\lambda$) agent was used in this task and the action-value function in expressed by the tabular function. In this experiment, the learning rate $\alpha$ was 0.1, the discount rate $\gamma$, 0.95 and the decay rate of the eligibility trace $\lambda$, 0.1. The agent follows the $\epsilon$-greedy policy in which the action is almost entirely selected by greedy action selection $a = \argmax_b Q(s, b)$ but, with a small probability of $\epsilon = 0.01$, the action is selected from the uniformly random distribution over the action set. The training procedure of the agent was as follows. An episode starts at the optimal battery level (that is, $B_t = 60$) with a random allocation of objects in the environment. At each time step, the alive flag is updated according to the temporal survival probability $P(A_{t+1}|s_t)$. If the agent receives $A_{t+1} = 0$ after the $t$-th update, the episode ends and the next episode starts.

\paragraph{Results}
\begin{figure}[!t]
\centering
\includegraphics[width=5cm]{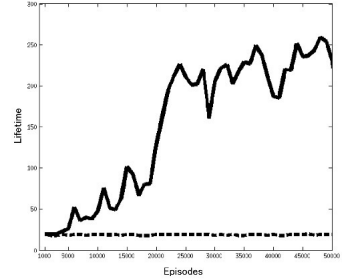}
\caption{The median of the survival time step along the episodes. The details are explained in the main text.}
\label{lifetime}
\end{figure}

\begin{figure}[!t]
\centering
\includegraphics[width=5.5cm]{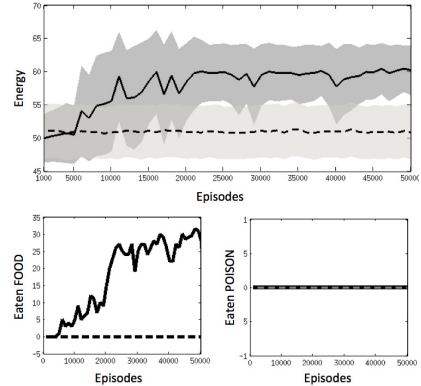}
\caption{Top: the evolution of the mean battery level and the standard deviation at death. Bottom left: median number of {\bf A} (food) eaten by the agent during the evaluation step. Bottom right: the median of the number of {\bf B} (poison) eaten by the agent during the evaluation step. Solid line: RL agent with survival reward settings. Dashed line: the agent with random action selection.}
\label{battery}
\end{figure}

Figure \ref{lifetime} shows the evolution of the median survival time along the number of episodes. Evaluation was done by freezing parameters of the agent every 1000 episodes, and the agent is tested in 1000 episodes without learning. The solid line represents the median of the survival time of the Sarsa($\lambda$) agent with the survival reward settings described above. The dashed line represents an agent that randomly and uniformly selects one action among the 5 actions. The growth of the lifetime clearly shows that the Sarsa($\lambda$) agent successfully learns the survival strategy during the process. The results of the random agent show that the environment has sufficient complexity that the random agent cannot stay alive longer than 25 time steps. Figure \ref{battery} shows the battery level of the agent at its death and the median amount of {\bf A} (left panel) and {\bf B} (right panel) consumed in the evaluation process after the corresponding episodes. From these figures, we can know that the battery level is successfully controlled around the desired level ($E = 60$) and that the amount of food eaten ({\bf A}) increased. On the other hand, the number of poisonous objects ({\bf B}) eaten is always zero. This result also supports the successful learning of the survival strategy by Sarsa($\lambda$) with only the survival rewards.

\section{Discussion}
In our approach, we introduced the first ``fundamental'' reward function of RL for the general survival problem, which so far has been  only heuristically defined in previous studies. The key is to soften the definition of the viability zone with the temporal survival probability, so that the reward function is simply the log of the temporal survival probability. Using this setting, the agents can learn the survival policy with respect to the maximization of the survival probability in the future. The source of the reward function, the temporal survival probability, has an explicit meaning and may be obtainable through the evolutionary process.
However, even though our reward setting is fundamental for survival, it may not be the ``optimal reward" in terms of learning efficiency for the survival policy. It is known that there are reward settings that have the same optimal policy but a different learning speed for the RL agent \cite{ng1999policy}.
And, recently, the pre-training approach for the RL has been successfully applied to the real-robot domain with direct visual image inputs \cite{lange2012autonomous}. Therefore, to speed up learning and achieve a robotic agent that fits the unknown-dynamic environment in its (sometimes physical) lifetime, the survival problem should be examined to determine how to equip an agent with moderate prior knowledge of the environment, including the reward function and the state transition dynamics.

The relationship between the planning problem and the inference problem is a hot topic in recent machine learning communities \cite{toussaint2006probabilistic,todorov2008general,kappen2012optimal}. Vlassis and Toussaint \cite{vlassis2009model} introduced the perspective that model-free reinforcement learning can be treated as an application of a stochastic EM algorithm to the maximization of the mixture likelihood $p(R=1; \pi)$. Our objective function (\ref{eq:suv}) and its lower bound were briefly introduced by Toussaint \cite{toussaint2009robot} in the context of a stochastic control problem. In this context, the goal of our study is the maximization of the joint probability (\ref{eq:suv}) from the beginning. Further, we demonstrated the relationships between its lower bound and the EM-based approach, including the POMDP case.

\section{Conclusion}
We have discussed the survival problem of the agent in the environment, and have shown that the survival problem can be reduced to an RL problem in (PO)MDPs with a specific reward function. Because of the popularity of the (PO)MDP assumptions in the research of autonomous agents, especially those concerning models of animals, this formulation may be seen as the basis of a truly autonomous agent for survival.


%



\section*{Acknowledgment}
I would like to thank Makoto Otsuka and Stephany Nix for helpful comments to improve the quality of this paper.



%
\bibliographystyle{IEEEtran}
\bibliography{reference}

\end{document}